\documentclass[acmsmall]{acmart}

\AtBeginDocument{%
  }
\setcopyright{acmcopyright}
\copyrightyear{2023}
\acmYear{2023}
\acmDOI{XXXXXXX.XXXXXXX}
\acmJournal{JACM}
\acmVolume{37}
\acmNumber{4}
\acmArticle{111}
\acmMonth{8}

\usepackage{amsmath,amsfonts}
\usepackage{algorithmic}
\usepackage{algorithm}
\usepackage{array}
\usepackage[caption=false,font=normalsize,labelfont=sf,textfont=sf]{subfig}
\usepackage{textcomp}
\usepackage{stfloats}
\usepackage{url}
\usepackage{verbatim}
\usepackage{graphicx}
\usepackage{hhline}
\usepackage{adjustbox}
\usepackage{float}
\usepackage{booktabs}
\usepackage{colortbl}
\usepackage{multirow}
\usepackage{tabu}
\usepackage{lscape}
\usepackage[normalem]{ulem}
\usepackage{wrapfig}
\usepackage{lettrine}
\usepackage{verbatim}
\usepackage{url}
\usepackage[utf8]{inputenc}
\usepackage{lscape}
\usepackage{comment, soul}
\usepackage{bm}
\usepackage{diagbox}
\usepackage{slashbox}
\usepackage{makecell}
  
\newcolumntype{C}{>{\centering\arraybackslash}X}
\usepackage{pifont}
\definecolor{redtable}{RGB}{255, 153, 153}
\definecolor{greentable}{RGB}{128, 255, 128}
\usepackage{bm}
\allowdisplaybreaks

\usepackage{hyperref}
\hypersetup{
    colorlinks=true,
    linkcolor=blue,
    filecolor=magenta,      
    urlcolor=cyan,
}

\newcommand\notsotiny{\@setfontsize\notsotiny{6}{7}}
\newcolumntype{x}[1]{>{\centering\arraybackslash\hspace{0pt}}p{#1}}

\usepackage{array, makecell}
\usepackage{adjustbox}

\usepackage{caption}

\usepackage{acronym}
\usepackage[normalem]{ulem}
\useunder{\uline}{\ul}{}

\newacro{dnn}[DNN]{deep neural network}
\newacro{cnn}[CNN]{convolution neural network}
\newacro{rnn}[RNN]{recurrent neural network}
\newacro{ann}[ANN]{artificial neural network}
\newacro{nlp}[NLP]{natural language processing}
\newacro{vvc}[VVC]{versatile video coding}
\newacro{hevc}[HEVC]{high-efficiency video coding}
\newacro{bpg}[BPG]{better portable graphics}
\newacro{vit}[ViT]{vision transformer}
\newacro{resblock}[ResBlock]{residual block}
\newacro{vae}[VAE]{variational autoencoder}
\newacro{vqvae}[VQ-VAE]{vector-quantized variational autoencoder}
\newacro{gdn}[GDN]{generalized divisive normalization}
\newacro{gelu}[GELU]{gaussian error linear unit}
\newacro{mha}[MHA]{multi-headed attention}
\newacro{gmm}[GMM]{gaussian mixture model}
\newacro{pq}[PQ]{product quantization}
\newacro{acnn}[acnn]{asymmetric convolutional neural network}
\newacro{mse}[MSE]{mean squared error}
\newacro{mim}[MIM]{masked image modeling}
\newacro{psnr}[PSNR]{peak signal-to-noise ratio}
\newacro{ms-ssim}[MS-SSIM]{multi-scale structural similarity index}
\newacro{flops}[FLOPs]{floating point operations per second}
\newacro{tfc}[TFC]{tensorflow compression}
\newacro{gpu}[GPU]{graphics processing unit}
\newacro{cpu}[CPU]{central processing unit}
\newacro{charm}[ChARM]{channel-wise autoregressive model}
\newacro{rpn}[RPN]{resize parameter network}
\newacro{stn}[STN]{spatial transformer network}
\newacro{nic}[NIC]{neural image compression}
\newacro{aict}[AICT]{adaptive image compression transformer}
\newacro{ict}[ICT]{image compression transformer}
\newacro{nic}[NIC]{neural image compression}
\newacro{npu}[NPU]{neural processing unit}
\newacro{mem}[MEM]{Multi-Reference Entropy Model}

\begin{document}

\title{Joint Hierarchical Priors and Adaptive Spatial Resolution for Efficient Neural Image Compression}

\author{Ahmed Ghorbel}
\email{ghorbel.ahmd@gmail.com}
\affiliation{%
  \institution{Univ. Rennes, INSA Rennes, CNRS, IETR - UMR 6164}
  \city{Rennes}
  \country{France}
}

\author{Wassim Hamidouche}
\email{Wassim.Hamidouche@tii.ae}
\affiliation{%
  \institution{Technology Innovation Institute, Masdar City, P.O Box 9639}
  \city{Abu Dhabi}
  \country{UAE}
}

\author{Luce Morin}
\email{luce.morin@insa-rennes.fr}
\affiliation{%
  \institution{Univ. Rennes, INSA Rennes, CNRS, IETR - UMR 6164}
  \city{Rennes}
  \country{France}
}

\renewcommand{\shortauthors}{Ghorbel et al.}

%
\begin{abstract}
%
Recently, the performance of \ac{nic} has steadily improved thanks to the last line of study, reaching or outperforming state-of-the-art conventional codecs.
%
Despite significant progress, current \ac{nic} methods still rely on ConvNet-based entropy coding, limited in modeling long-range dependencies due to their local connectivity and the increasing number of architectural biases and priors, resulting in complex underperforming models with high decoding latency.
%
Motivated by the efficiency investigation of the Tranformer-based transform coding framework, namely SwinT-ChARM, we propose to enhance the latter, as first, with a more straightforward yet effective Tranformer-based channel-wise autoregressive prior model, resulting in an absolute \ac{ict}. Through the proposed \ac{ict}, we can capture both global and local contexts from the latent representation and better parameterize the distribution of the quantized latents. Further, we leverage a learnable scaling module with a sandwich ConvNeXt-based pre-/post-processor to accurately extract more compact latent codes while reconstructing higher-quality images.
%
Extensive experimental results on benchmark datasets showed that the proposed framework significantly improves the trade-off between coding efficiency and decoder complexity over the \ac{vvc} reference encoder (VTM-18.0) and the neural codec SwinT-ChARM.
%
Moreover, we provide model scaling studies to verify the computational efficiency of our approach and conduct several objective and subjective analyses to bring to the fore the performance gap between the \ac{aict} and the neural codec SwinT-ChARM.
%
All materials, including the source code of \ac{aict}, will be made publicly accessible upon acceptance for reproducible research.
\end{abstract}



\received{15 January 2023}

\maketitle

\acresetall
%
\section{Introduction}
\label{intro}
%
Visual information is crucial in human development, communication, and engagement, and its compression is necessary for effective storage and transmission over constrained wireless and wireline channels. Thus, thinking about new enhanced lossy image compression solutions is a goldmine for scientific research. The goal is to reduce an image file size by permanently removing redundant data and less critical information, particularly high frequencies, to obtain the most compact bit-stream representation while preserving a certain level of visual fidelity. Despite this, optimizing the rate-distortion tradeoff involves a fundamental objective for achieving a high compression ratio and low distortion.
\begin{figure}[!hbt]
\centering
\includegraphics[width=1\textwidth]{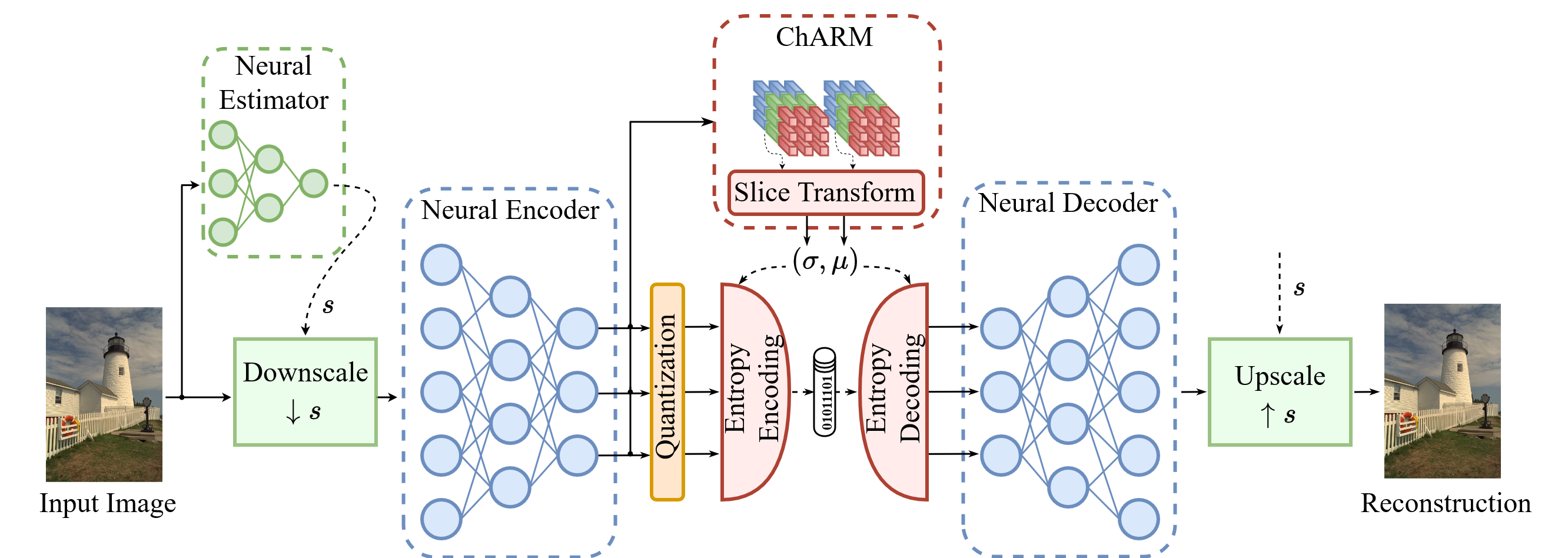}
    \caption{A high-level diagram of the proposed AICT solution. ChARM refers to the Transformer-based channel-wise autoregressive prior model, and $s$ represents the resizing parameter predicted by the neural estimator $(s \in \mathbb{R} \cap [0, 1])$.}
\label{hldiag}
\end{figure}
%

Conventional image and video compression standards, including JPEG~\cite{125072}, JPEG2000~\cite{899217}, H.265/\ac{hevc}~\cite{6316136}, and H.266/\ac{vvc}~\cite{9503377}, rely on hand-crafted creativity within a block-based encoder/decoder diagram \cite{10.1145/248621.248622}. In addition, recent conventional codecs \cite{10.1145/3471905,10.1145/3408320,10.1145/3631710,10.1145/3633459} employ intra-prediction, fixed transform matrices, quantization, context-adaptive arithmetic encoders, and various in-loop filters to reduce spatial and statistical redundancies and alleviate coding artifacts. However, it has taken several years to standardize a conventional codec. Moreover, existing image compression standards are not anticipated to be an ideal and global solution for all types of image content due to the rapid development of new image formats and the growth of high-resolution mobile devices. 

On the other hand, with recent advancements in machine learning and artificial intelligence, new \ac{nic} schemes \cite{8693636} have emerged as a promising alternative to traditional compression methods. \ac{nic} consists of three modular parts: transform, quantization, and entropy coding. Each of these components can be represented as follows: i) autoencoders as flexible nonlinear transforms where the encoder (i.e., analysis transform) extracts a latent representation from an input image and the decoder (i.e., synthesis transform) reconstructs the image from the decoded latent, ii) differentiable quantization that quantizes the encoded latent iii) \textcolor{black}{deep prior model} estimating the conditional probability distribution of the quantized latent to reduce the rate. Further, these three components are jointly optimized in end-to-end training by minimizing the distortion loss between the original image and its reconstruction, and the rate needed to transmit the bit-stream of latent representation.

Recently, we have seen a significant surge of deep learning-based lines of study exploring the potential of \acp{ann} to develop various \ac{nic} frameworks, reaching or even outperforming state-of-the-art conventional codecs. Some of these previous works leverage hyperprior-related side information \cite{balle2018variational,hu2020coarse} to capture short-range spatial dependencies or additional context model \cite{minnen2018joint,lee2019extended}, and others use non-local mechanism \cite{cheng2020learned,li2021learning,qian2021learning,chen2021end} to model long-range spatial dependencies. For example, Mentzer et al. \cite{mentzer2020high} proposed a generative compression method achieving high-quality reconstructions. In contrast, Minnen et al.  \cite{minnen2020channel} introduced channel-conditioning and latent residual prediction, taking advantage of an entropy-constrained model that uses both forward and backward adaptations.
Current research trends has focused on attention-guided compressive transform, as Zhu et al. \cite{zhu2021transformer} replaced the ConvNet-based transform coding in the Minnen et al. \cite{minnen2020channel} architecture with a Transformer-based nonlinear transform. Later, Zou et et al. \cite{zou2022devil} combined the local-aware attention mechanism with the global-related feature learning and proposed a window-based attention module.
An additional series of efforts have addressed new entropy coding methods, as Zhu et al. \cite{zhu2022unified} proposed a probabilistic vector quantization with cascaded estimation to estimate pairs of mean and covariance under a multi-codebooks structure. Guo et al.~\cite{9455349} introduced the concept of separate entropy coding by dividing the latent representation into two channel groups, and proposed a causal context model that makes use of cross-channel redundancies to generate highly informative adjacent contexts. Further, Kim et al. \cite{kim2022joint} exploited the joint global and local hyperprior information in a content-dependent manner using an attention mechanism. He et al. \cite{he2022elic} adopted stacked residual blocks as nonlinear transform and multi-dimension entropy estimation model.
More recently, El-Nouby et al. \cite{el-nouby2023image} replaced the vanilla vector quantizer with \ac{pq} \cite{jegou2008accurate} in a compression system derived from \ac{vqvae} \cite{van2016conditional} offering a large set of rate-distortion points and then introduced a novel \ac{mim} conditional entropy model that improves entropy coding by modeling the co-dependencies of the quantized latent codes. Also, Muckley et al. \cite{muckley2023improving} introduced a new adversarial discriminator based on \ac{vqvae} that optimizes likelihood functions in the neighborhood of local images under the mean-scale hyperprior Minnen et al. \cite{minnen2018joint} architecture. Additionally, Chen et al. \cite{10.1145/3617733.3617746} formulated an improved method called Adaptive \ac{vqvae} to compactly represent the latent space of convolutional neural network. Further, Xue et al. \cite{10.1145/3595916.3626372} proposed an exponential R-$\lambda$ model for accurate bitrate estimation, along with a multi-layer feature modulation mechanism in the compression network to ensure monotonic bitrate variation with $\lambda$. Moreover, Lv et al. \cite{10.1145/3581783.3612187} proposed a low-rank adaptation approach, which updates decoder weights using low-rank decomposition matrices at inference time, and a dynamic gating network that learns the optimal number and positions of adaptation. Jiang et al. \cite{10.1145/3581783.3611694} introduced the \ac{mem} and its advanced version, \ac{mem}+, designed to capture diverse correlations in the latent representations by employing attention map and enhanced checkerboard context for capturing both local and global spatial contexts.

Other interesting attempts \cite{dupont2021coin,strumpler2022implicit}, known as coordinate-based or implicit neural representations, have shown good ability to represent, generate, and manipulate various data types, particularly in \ac{nic} by training image-specific networks that map image coordinates to RGB values, and compressing the image-specific parameters. On the other hand, Wu et al. \cite{9568930} proposed a learned block-based hybrid image compression method, which introduces a contextual prediction module to utilize the relationship between adjacent blocks, and propose a boundary-aware post-processing module to remove the block artifacts.

Through these numerous pioneering works, we can estimate the importance of \ac{nic} in the research field and the industry. Thus, identifying the main open challenges in this area is crucial. The first one is to discern the most relevant information necessary for the reconstruction, knowing that information overlooked during encoding is usually lost and unrecoverable for decoding. The second challenge is to enhance the trade-off between coding efficiency and decoding latency. While the existing approaches improve the transform and entropy coding accuracy, they still need to improve the decoding latency and reduce the model complexity, leading to an ineffective real-world deployment. 
To tackle those challenges, we propose a nonlinear transform coding and channel-wise autoregressive entropy coding built on Swin Transformer~\cite{liu2021swin} blocks and paired with a neural scaling network, namely \ac{aict}. Fig.~\ref{hldiag} portrays a high-level diagram to provide a more comprehensive overview of the proposed framework.

The contributions of this paper are summarized as follows:
\begin{itemize}
\item We propose the \ac{ict}, a nonlinear transform coding and spatio-channel autoregressive entropy coding. These modules are based on Swin Transformer blocks for effective latent decorrelation and a more flexible receptive field to adapt to contexts requiring short/long-range information.
\item We propose the \ac{aict} model that adopts a scale adaptation module as a sandwich processor to enhance compression efficiency. This module consists of a neural scaling network, and ConvNeXt-based \cite{liu2022convnet} pre-/post-processor to optimize differentiable resizing layers jointly with a content-dependent resize factor estimator.
\item We conduct extensive experiments on four widely-used benchmark datasets to explore possible coding gain sources and demonstrate the effectiveness of \ac{aict}. In addition, we carried out a model scaling analysis and an ablation study to substantiate our architectural decisions.
\end{itemize}
The experimental results reveal the impact of the spatio-channel entropy coding, the sandwich scale adaptation component, and the joint global structure and local texture learned by the attention units through the nonlinear transform coding. These experiments show that the proposed  \ac{ict} and \ac{aict}  achieve respectively -4.65\%  and -5.11\% BD-rate (\acs{psnr}) reduction over VTM-18.0 while considerably reducing the decoding latency, outperforming conventional and neural codecs in the trade-off between coding efficiency and decoding complexity.

The rest of this paper is organized as follows. First, Section~\ref{sec:background} briefly describes the background and related works. Then, Section~\ref{methd} presents our overall framework along with a detailed description of the proposed architecture. Further, we devote Section~\ref{result} to present and analyze the experimental results. Finally, Section~\ref{concl} concludes the paper.

%
\section{Background and Related Works}
\label{sec:background}
Over the past years, research has renewed interest in modeling image compression as a learning problem, giving a series of pioneering works \cite{balle2018variational,minnen2018joint,lee2019extended,choi2019variable,li2020learning,9204799,cheng2020learned,hu2020coarse,minnen2020channel} that have contributed to a universal fashion effect, and have achieved great success, augmented by the efficient connection to variational learning \cite{gregor2016towards,frey1997bayesian,alemi2018fixing}. In the early stage, some of these methods adopted ConvNets and activation layers coupled with \ac{gdn} layers to perform non-linear transform coding over a \ac{vae} architecture. This framework creates a compact representation of the image by encoding them to a latent representation. The compressive transform squeezes out the redundancy in the image with dimensional reduction and entropy constraints. Following that, some studies focus on developing network architectures that extract compact and efficient latent representation while providing higher-quality image reconstruction.

This section reviews relevant \ac{nic} techniques, including works related to our research, while focusing on the following aspects. First, we briefly present the autoregressive context related works. Then, we describe the end-to-end \ac{nic} methods that have recently emerged, including attention-guided and Transformer-based coding. Finally, we introduce adaptive downsampling within the context of neural coding. 

\subsection{Autoregressive Context}
%
Following the success of autoregressive priors in probabilistic generative models, Minnen et al. \cite{minnen2018joint} was the first to introduce autoregressive and hierarchical priors within the variational image compression framework, featuring a mean-scale hyperprior. An additional context model is added to boost the rate-distortion performance. Although the combined model demonstrated superior rate-distortion performance compared to neural codecs, it came with a notable computational cost.
%
Later, Cheng et al.~\cite{cheng2020learned} proposed the first model achieving competitive coding performance with \ac{vvc}, using a context model in an autoregressive manner. They improved the entropy model by using a discretized K-component \ac{gmm}.
%
In addition, Minnen et al. \cite{minnen2020channel} estimated the latent distribution's mean and standard deviation in a channel-wise manner and incorporated an autoregressive context model to condition the already-decoded latent slices and the latent rounding residual on the hyperprior to further reduce the spatial redundancy between adjacent pixels.
%
Finally, He et al. \cite{he2021checkerboard} proposed a parallelizable spatial context model based on the checkerboard-shaped convolution that allows parallel-friendly decoding implementation, thus increasing the decoding speed.

\subsection{Attention-Guided Coding}
Attention mechanism was popularized in \ac{nlp} \cite{luong2015effective,vaswani2017attention}. It can be described as a mapping strategy that queries a set of key-value pairs to an output. For example, Vaswani et al.~\cite{vaswani2017attention} have proposed \ac{mha} methods in which machine translation is frequently used. For low-level vision tasks \cite{mentzer2018conditional,zhang2018residual,li2020learning}, spatially adaptive feature activation is made possible by the attention mechanism, focusing on more complex areas, like rich textures, saliency, etc. 
%
In image compression, quantized attention masks are used for adaptive bit allocation, e.g., Li et al.~\cite{li2020learning} used a trimmed convolutional network to predict the conditional probability of quantized codes, Mentzer et al.~\cite{mentzer2018conditional} relied on a 3D-\ac{cnn}-based context model to learn a conditional probability model of the latent distribution. Later, Cheng et al.~\cite{cheng2020learned} inserted a simplified attention module (without the non-local block) into the analysis and synthesis transforms to pay more attention to complex regions. More recently, Zou et al.~\cite{zou2022devil} combined the local-aware attention mechanism with the global-related feature learning within an effective window-based local attention block, which can be used as a specific component to enhance ConvNet and Transformer models.
Guo et al.~\cite{9455349} adopted a powerful group-separated attention module to strengthen the non-linear transform networks. Further, Tang et al.~\cite{9858899} integrated graph attention and \ac{acnn} for end-to-end image compression, to effectively capture long-range dependencies and emphasize local key features, while ensuring efficient information flow and reasonable bit allocation.

\subsection{Transformer-based Coding} 
%
Recently, Transformers have been increasingly used in neural codecs. They exempt convolution operators entirely and rely on attention mechanisms to capture the interactions between inputs, regardless of their relative position, thus allowing the network to focus more on pertinent input data elements. Qian et al.~\cite{qian2022entroformer} replaced the autoregressive hyperprior \cite{minnen2018joint} with a self-attention stack and introduced a novel Transformer-based entropy model, where the Transformer's self-attention is used to relate different positions of a single latent for computing the latent representation. Zhu et al. \cite{zhu2021transformer} replaced all convolutions in the standard approach \cite{minnen2020channel,balle2018variational} with Swin Transformer~\cite{liu2021swin} blocks, leading to a more flexible receptive field to adapt tasks requiring both short/long-range information, and better progressive decoding of latent. Apart from their effective window-based local attention block, Zou et al.  \cite{zou2022devil} proposed a novel symmetrical Transformer (STF) framework with absolute Transformer blocks for transform coding combined with a \ac{charm} prior. Inspired by the adaptive characteristics of the Transformers, Koyuncu et al. \cite{koyuncu2022contextformer} proposed a Transformer-based context model, which generalizes the de facto standard attention mechanism to spatio-channel attention.

\subsection{Adaptive Downsampling}
Learning sampling techniques were first developed for image classification to improve image-level prediction while minimizing computation costs. \acp{stn} \cite{jaderberg2015spatial} introduced a layer that estimates a parametrized affine, projective, and splines transformation from an input image to recover data distortions and thereby improve image classification accuracy. Recasens et al. \cite{recasens2018learning} suggested that when downsampling an input image for classification, salient regions should be "zoomed-in" to learn a saliency-based network jointly. Talebi et al. \cite{talebi2021learning} jointly optimize pixel value interpolated at each fixed downsampling location for classification. Marin et al. \cite{marin2019efficient} recently argued that a better downsampling scheme should sample pixels more densely near object boundaries, and introduced a strategy that adapts the sampling locations based on the output of a separate edge-detection model. Further, Jin et al. \cite{jin2022learning} introduced a deformation module and a learnable downsampling operation, which can be optimized with the given segmentation model in an end-to-end fashion.

In the context of \ac{nic}, Chen et al. \cite{chen2022estimating} proposed a straightforward learned downsampling module that can be jointly optimized with any \ac{nic} kernels in an end-to-end fashion. Based on the \ac{stn} \cite{jaderberg2015spatial}, a learned resize parameter is used in a bilinear warping layer to generate a sampling grid, where the input should be sampled to produce the resampled output. They also include an additional warping layer necessary for an inverse transformation to maintain the same resolution as the input image.

%
\begin{figure*}[htb]
\centering
\includegraphics[width=1\textwidth]{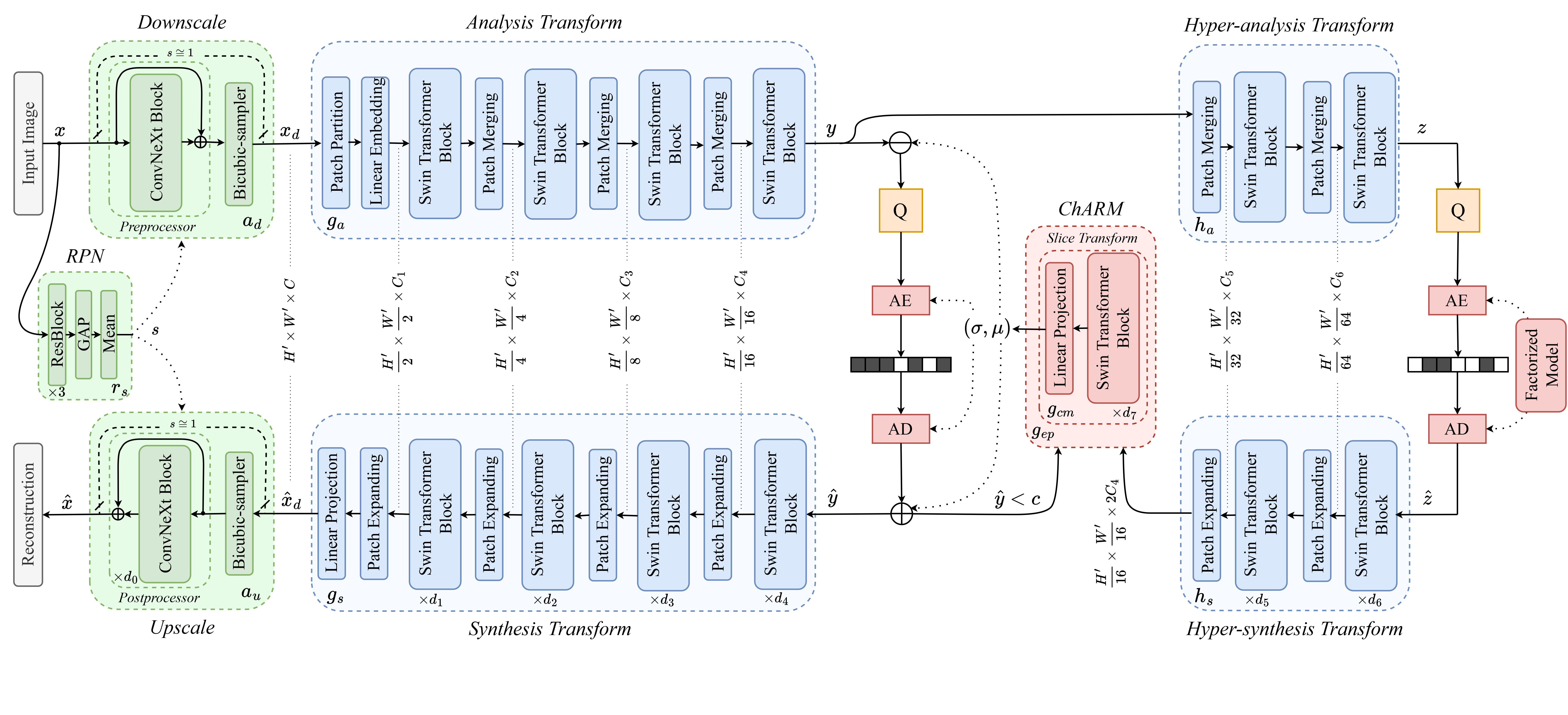}
\caption{Overall AICT Framework. We illustrate the image compression diagram of our AICT with hyperprior and Swin Transformer based \ac{charm}, and scale adaptation module. The \ac{rpn}, ConvNeXt block, and Swin Transformer block architectures are respectively detailed in (a), (c), and (d) Fig.~\ref{blocks}.}
\label{ov_framework}
\end{figure*}
\section{Proposed AICT framework}
\label{methd}
In this section, we first formulate the \ac{nic} problem. Next, we introduce the design methodology for the overall \ac{aict} architecture, followed by the description of each component individually.

\subsection{Problem Formulation}
%
\textcolor{black}{The primary challenges addressed in this work are twofold. Firstly, we aim to identify and prioritize the most pertinent information required for accurate reconstruction. It is crucial to acknowledge that any information overlooked during the encoding phase is typically lost and irretrievable during decoding. Secondly, we endeavor to optimize the delicate balance between coding efficiency and decoding latency. While existing approaches have made strides in improving the accuracy of transform and entropy coding, there remains a pressing need to mitigate decoding latency and streamline model complexity for practical real-world deployment. To tackle these challenges, we introduce a novel approach, denoted as \ac{aict}. \ac{aict} leverages nonlinear transform coding and channel-wise autoregressive entropy coding techniques, building upon Swin Transformer blocks and incorporating a neural scaling network. In the context of describing these contributions, it is imperative to establish a clear problem formulation, which we delve into in the following section.
}
The objective of \ac{nic} is to minimize the distortion between the original image and its reconstruction under a specific distortion-controlling hyperparameter. For an input image $\bm{x}$, the analysis transform $g_{a}$, with parameter $\phi_{g}$, removes the image spatial redundancies and generates the latent representation $\bm{y}$. Then, this latent is quantized to the discrete code $\hat{\bm{y}}$ using the quantization operator $\lceil$.$\rfloor$, from which a synthesis transform $g_{s}$, with parameter $\theta_{g}$, reconstructs the image denoted by $\hat{\bm{x}}$. The overall process can be formulated as follows:
\begin{equation}
\begin{aligned}
\bm{y} &= g_{a}( \bm{x} \mid \phi_{g}), \\
\hat{\bm{y}} &= \lceil \bm{y} \rfloor, \\
\hat{\bm{x}} &= g_{s}(\hat{\bm{y}} \mid \theta_{g}).
\end{aligned}
\end{equation}

A hyperprior model composed of a hyper-analysis and hyper-synthesis transforms $(h_{a}, h_{s})$ with parameters $(\phi_{h}, \theta_{h})$ is usually used to reduce the statistical redundancy among latent variables. In particular, this hyperprior model assigns a few extra bits as side information to transmit some spatial structure information and helps to learn an accurate entropy model. The generated hyper-latent representation $\bm{z}$ is quantized to the discrete
code $\hat{\bm{z}}$ using the quantization operator $\lceil$.$\rfloor$. The hyperprior generation can be summarized as follows:
\begin{equation}
\begin{aligned}
\bm{z} &= h_{a}(\bm{y} \mid \phi_{h}), \\
\hat{\bm{z}} &= \lceil \bm{z} \rfloor, \\
p_{\hat{\bm{y}} \mid \hat{\bm{z}}}(\hat{\bm{y}} \mid \hat{\bm{z}}) & \leftarrow h_{s}(\hat{\bm{z}} \mid \theta_{h}).
\end{aligned}
\end{equation}
%
Further, considering a context model $g_{cm}$ with parameter $\psi_{cm}$, and a parameter inference network $g_{ep}$ with parameter $\psi_{ep}$ which estimates, from the latent $\hat{\bm{y}}$, the location and scale parameters $\Phi=(\mu, \sigma)$ of the entropy model. 
The parameter prediction for $i$-th representation $\hat{\bm{y}}_i$ is expressed as follows:
\begin{equation}
\begin{aligned}
\Phi_{i}=g_{ep}(h_{s}(\hat{\bm{z}}), g_{cm}(\hat{\bm{y}}_{<i} \mid \psi_{cm})  \mid \psi_{ep}),
\end{aligned}
\end{equation}
where $\Phi_{i}=(\mu_{i}, \sigma_{i})$ is used to jointly predict entropy parameters, and $\hat{\bm{y}}_{<i} =\{\hat{\bm{y}}_1, \ldots, \hat{\bm{y}}_{i-1}\}$ is the observable neighbors of each symbol vector $\hat{\bm{y}}_i$ at the $i$-th location.
\begin{equation}
\begin{aligned}
p_{\hat{\bm{y}}_{i} \mid \hat{\bm{z}}}(\hat{\bm{y}}_{i} \mid \hat{\bm{z}})=\sum_{0<k<K} \pi_{i}^{k}[\mathcal{N}_{(\mu_{i}^{k}, \sigma_{i}^{2k})} * \mathcal{U}_{(-\frac{1}{2}, \frac{1}{2})}](\hat{\bm{y}}_{i}),
\end{aligned}
\end{equation}
where $\mathrm{K}$ groups of entropy parameters $(\pi^{k}, \mu^{k}, \sigma^{k})$ are calculated by $g_{\text{ep}}$, $\mathcal{N}_{ (\mu, \sigma^{2}) }$ represents the mean and scale Gaussian distribution, and $\mathcal{U}_{(-\frac{1}{2}, \frac{1}{2})}$ denotes the uniform noise.
%

%
Both transform and quantization introduce distortion $D = MSE(\bm{x}, \hat{\bm{x}})$ for \ac{mse} optimization that measures the reconstruction quality with an estimated bitrate $R$, corresponding to the expected rate of the quantized latent and hyper-latent, as described below:
\begin{equation}
R =  \mathbb{E} \left [-\log _{2}(p_{\hat{\bm{y}} \mid \hat{\bm{z}}}(\hat{\bm{y}} \mid \hat{\bm{z}})) -\log _2(p_{\hat{\bm{z}}}(\hat{\bm{z}}))\right ].
\end{equation}

In the case of adaptive resolution (i.e., \acs{aict}), we consider the \ac{rpn}, the downscale, and the upscale modules as $(r_{s}, a_{d}, a_{u})$ with parameters $(\omega_{r}, \omega_{d}, \omega_{u})$, respectively. The generation process of $\bm{x}_{d}$ and $\hat{\bm{x}}$ is described as follows:
\begin{equation}
\begin{aligned}
s &= r_{s}(\bm{x} \mid \omega_{r}), \\
\bm{x}_{d} &= a_{d}(\bm{x}, {s} \mid \omega_{d}), \\
\hat{\bm{x}} & =  a_{u}(\hat{\bm{x}}_{d}, {s} \mid \omega_{u}).
\end{aligned}
\end{equation}

Representing $(g_{a},g_{s})$, $(h_{a},h_{s})$, $(g_{cm},g_{ep})$, and $(r_{s}, a_{d}, a_{u})$ by \acp{dnn} enables jointly optimizing the end-to-end model by minimizing the rate-distortion trade-off $\mathcal{L}$, giving a rate-controlling hyperparameter $\lambda$. This optimization problem can be expressed as follows: 
%
%
\begin{equation}
\label{lossfct}
\begin{split}
{\arg \min} \mathcal{L} (\bm{x},  \bm{\hat{x}})
& = {\arg \min}D (\bm{x},  \bm{\hat{x}})+\lambda R, \\
& = {\arg \min}||\bm{x}- \bm{\hat{x}}||^2_2 + \lambda ( \underbrace{ \mathbb{H}(\bm{\hat{y}}) + \mathbb{H} (\bm{\hat{z}}) }_R ),
\end{split}
\end{equation}
where $\mathbb{H}$ stands for the cross entropy.

Finally, we recall that training the model with the gradient descent method requires substituting the quantization with additive uniform noise \cite{balle2016end}, preventing the gradient from vanishing at the quantization. We follow this method in this paper, where the noisy representations of the latent are used to compute the rate during the training phase.  

\subsection{Overall Architecture} 
The overall pipeline of the proposed solution is illustrated in Fig.~\ref{ov_framework}. The framework includes three modular parts. First, the scale adaptation module, composed of a tiny \acf{rpn}~\cite{chen2022estimating}, a ConvNeXt-based pre-/post-processor, and a bicubic interpolation filter. Second, the analysis/synthesis transform $(g_{a},g_{s})$ of our design consists of a combination of patch merging/expanding layers and Swin Transformer~\cite{liu2021swin} blocks. The architectures of hyper-transforms $(h_{a},h_{s})$ are similar to $(g_{a},g_{s})$ with different stages and configurations. Then, a Transformer-based slice transform inside a \ac{charm} is used to estimate the distribution parameters of the quantized latent. Finally, the resulting discrete-valued data $(\hat{\bm{y}}, \hat{\bm{z}})$ are encoded into bit-streams with an arithmetic encoder.

\subsection{Scale Adaptation Module}
\begin{figure}[t]
\centering
\includegraphics[width=1\textwidth]{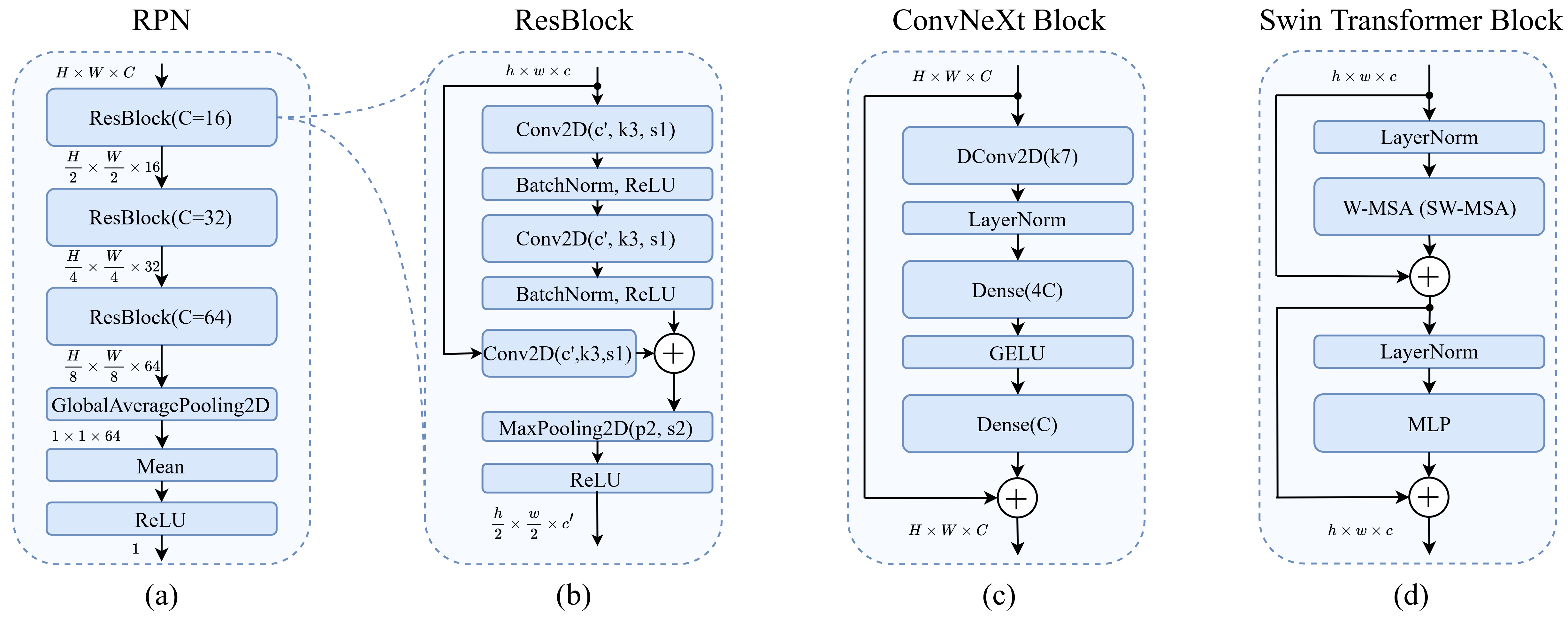}
\vspace{-2mm}
\caption{Detailed description of block architectures: (a) \ac{rpn}, (b) ResBlock, (c) ConvNeXt Block, and (d) Swin Transformer Block. \textcolor{black}{DConv2D(.) stands for depthwise 2D convolution, LayerNorm for the layer normalization, Dense(.) for the densely-connected neural network layer, and \ac{gelu} for the activation.}}
\vspace{-4mm}
\label{blocks}
\end{figure}

Given a source image $\bm{x} \in \mathbb{R}^{H \times W \times C}$, we first determine an adaptive spatial resize factor $s \in \mathbb{R} \cap [0, 1]$ estimated by the \ac{rpn} module, which consists of three stages of \acp{resblock}. Indeed, the estimated resize parameter $s$ is used to create a sampling grid $\tau_{M}$ following the convention \acp{stn}, and used to adaptively down-scale $\bm{x}$ into $\bm{x}_{d} \in \mathbb{R}^{H' \times W' \times C}$  through the bicubic interpolation, with $H'=s \, H$ and $W'= s \, W$. The latter (i.e., $\bm{x}_{d}$) is then encoded and decoded with the proposed \ac{ict}. Finally, the decoded image $\hat{\bm{x}}_{d} \in \mathbb{R}^{H' \times W' \times C}$ is up-scaled to the original resolution $\hat{\bm{x}} \in \mathbb{R}^{H \times W \times C}$ using the same, initially estimated, resize parameter $s$.
The parameterization of each layer is detailed in the \ac{rpn} and \ac{resblock} diagrams of Fig.~\ref{blocks} (a) and (b), respectively. In addition, a learnable depth-wise pre-/post-processor is placed before/after the bicubic sampler to mitigate the information loss introduced by down/up-scaling, allowing the retention of information. This neural pre-/post-processing method consists of concatenation between the input and the output of three successive ConvNeXt~\cite{liu2022convnet} blocks, using depth-wise convolutions with large kernel sizes to obtain efficient receptive fields.
%
Globally, the ConvNeXt block incorporates a series of architectural choices from a Swin Transformer while maintaining the network's simplicity as a standard ConvNet without introducing any attention-based module. These design decisions can be summarized as follows: macro design, ResNeXt's grouped convolution, inverted bottleneck, large kernel size, and various layer-wise micro designs \cite{liu2022convnet}. In Fig.~\ref{blocks} (c), we illustrate the ConvNeXt block, where the DConv2D(.) refers to the depthwise 2D convolution, LayerNorm for the layer normalization, Dense(.) for the densely-connected neural network layer, and \ac{gelu} for the activation function. Finally, it is essential to note that we propose to skip the scale adaptation module for a better complexity-efficient design when the predicted scale does not change the input resolution, i.e., $s \cong 1$. The overhead to store and transmit the scale parameter $s$ can be ignored, given the large bitstream size of the image.

\subsection{Transformer-based Analysis/Synthesis Transform}
The analysis transform $g_a$ contains four stages of patch merging layer and Swin Transformer block to obtain a more compact low-dimensional latent representation $\bm{y}$. In order to consciously and subtly balance the importance of feature compression through the end-to-end learning framework, we used two additional stages of patch merging layer and Swin Transformer block in the hyper-analysis transform to produce the hyperprior latent representation $\bm{z}$.
During training, both latents $\bm{y}$ and $\bm{z}$ are quantized using a rounding function to produce $\hat{\bm{y}}$ and $\hat{\bm{z}}$, respectively. During inference, both latents $\bm{y}$ and $\bm{z}$ are first quantized using the same rounding function as training and then compressed using probability tables.
The quantized latent variables $\hat{\bm{y}}$ and $\hat{\bm{z}}$ are then entropy coded regarding an indexed entropy model for a location-scale family of random variables parameterized by the output of the \ac{charm}, and a batched entropy model for continuous random variables, respectively, to obtain the bit-streams. Finally, quantized latents $\hat{\bm{y}}$ and $\hat{\bm{z}}$ feed the synthesis and hyper-synthesis transforms, respectively, to generate the reconstructed image. The decoder schemes are symmetric to those of the encoder, with patch-merging layers replaced by patch-expanding layers.

The Swin Transformer block architecture, depicted in Fig.~\ref{blocks} (d), is a variant of the \ac{vit} that has recently gained attention due to its superior performance on a range of computer vision tasks. Therefore, it is essential to highlight the unique features and advantages to motivate the choice of the Swin Transformer over other \ac{vit} variants.
One key advantage of the Swin Transformer is its hierarchical design, which enables it to process images of various resolutions efficiently. Unlike other \ac{vit} variants, Swin Transformer divides the image into smaller patches at multiple scales, allowing it to capture both local and global information. This hierarchical design has been shown to be particularly effective for large-scale vision tasks.
Another advantage of the Swin Transformer is its ability to incorporate spatial information into its attention mechanism. Swin Transformer introduces a novel shifted window attention mechanism, which aggregates information from neighboring patches in a structured way, allowing it to capture spatial relationships between image features,  leading to linear complexity w.r.t. the input resolution. This attention mechanism has been shown to outperform the standard \ac{vit} attention mechanism, whose complexity is quadratic, on a range of benchmarks.
Overall, Swin Transformer's efficiency and superior performance make it a promising architecture for \ac{nic}. In addition, its ability to capture both global and local features efficiently, and its adaptability to different image resolutions, make it a strong contender among other transformer-based architectures.

\subsection{Transformer-based Slice Transform}
%
\textcolor{black}{In the realm of neural image compression, incorporating spatio-channel dependencies into entropy modeling is crucial. These dependencies, often termed spatial and channel or spatial-channel dependencies in prior literature~\cite{he2022elic,chen2019neural,10.1145/3581783.3611694}, capture the intricate relationships among neighboring pixels within quantized latent features across different channels. Recognizing these spatio-channel dependencies in the quantized latent representation is essential for eliminating redundancy along spatial and channel axes, ultimately enhancing compression efficiency while maintaining perceptual quality.
%
Aligned with this objective, we introduce the concept of separate entropy coding by partitioning the hyperprior latent representation into channel groups, as opposed to the conventional serial-decoded approach, to achieve more effective context modeling. Subsequently, we propose a tiny Transformer-based spatial context model that leverages cross-channel redundancies to generate highly informative adjacent contexts from the hyperprior latent slices, combined with the already-decoded latent slices. Consequently, our approach introduces a multidimensional entropy estimation model known as spatio-channel entropy modeling, which proves to be both fast and effective in reducing bitrate. According to this method, each latent element is conditioned on adjacent decoded elements that are spatio-channel neighbors, effectively eliminating redundancy along spatial and channel axes.}
\begin{figure}[htbp!]
  \centering
    \includegraphics[width=0.6\linewidth]{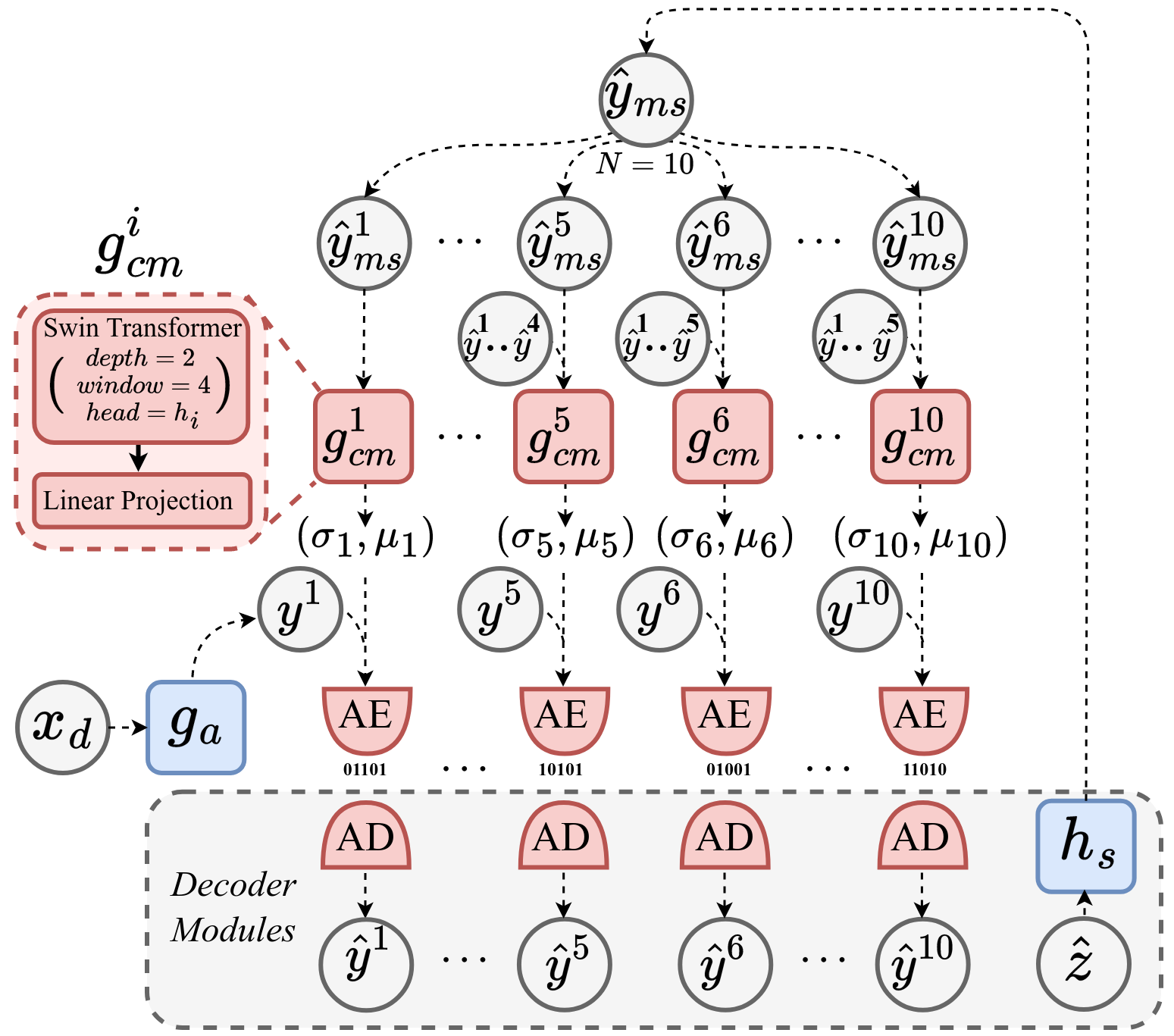}
  \caption{{\color{black} Spatio-channel entropy coding. $g_a$, $h_s$, $g^{i}_{cm}$, AE, and AD stand for analysis and hyper-synthesis transforms, the $i^{th}$ context model, and arithmetic encoder/decoder, respectively. $\{y^1,\dots,y^s\}$ stands for the already-decoded latent slices, where $\{s \in \mathbb{N} \mid 1 \leq s \leq 5 \}$ is the number of supported slices.}}
  \label{scec}
\end{figure}

%
\textcolor{black}{Fig.~\ref{scec} shows our spatio-channel entropy coding. We apply a tiny spatial context model $g^{i}_{cm}$ to exploit the spatio-channel correlations per $i^{th}$ grouped hyperprior channels $\hat{y}^{i}_{ms}$ combined with the already-decoded latent slices $\{y^1,\dots,y^s\}$, where $\{s \in \mathbb{N} \mid 1 \leq s \leq 5 \}$ stands for the number of supported slices. This process enhances the accuracy of entropy parameter estimation, thereby optimizing the overall efficiency of entropy coding.}
\textcolor{black}{As a side effect, it also results in faster decoding speed, thanks to the parallelization capabilities of the Swin Transformer on \ac{gpu} \cite{liu2021swin}. In contrast to ConvNets that rely on convolution in place of general matrix multiplication and are susceptible to communication overhead when parallelizing across multiple \acp{gpu}, Swin Transformers exhibit better parallelizability on \acp{gpu}. This is attributed to their hierarchical attention mechanism, which processes attention in a windowed manner, reducing global attention complexity and enabling efficient self-attention parallelization. Additionally, Swin Transformers leverage multi-head parallelism and tokenization strategies to maximize \ac{gpu} utilization. These features make them a prime choice for a wide range of computer vision tasks that require \ac{gpu} acceleration.}
%
The tiny slice transform consists of two successive Swin Transformer blocks with an additional learnable linear projection layer, used to get a representative latent slices concatenation. This \ac{charm} estimates the distribution $p_{\hat{\bm{y}}} (\hat{\bm{y}} | \hat{\bm{z}})$ with both the mean and standard deviation of each quantized latent slice and incorporates an autoregressive context model to condition the already-decoded latent slices and further reduce the spatial redundancy between adjacent pixels.

%
\section{Experimental Results}
\label{result}
In this section, we first describe the experimental setup, including the used datasets, the baselines against which we compared, and the implementation details. Then, we assess the compression efficiency of our method with a rate-distortion comparison and compute the average bitrate savings on four commonly-used evaluation datasets. We further elaborate a model scaling study to consistently examine the effectiveness of our proposed method against pioneering ones. Additionally, we perform a resize parameter analysis to show the variations of the predicted parameter $s$. Finally, we conduct a latent analysis, an ablation study, and a qualitative analysis to highlight the impact of our architectural choices.

\begin{table}[t]
\centering
\caption{Architecture configuration.}\label{config}
\begin{tabular}{@{}l|cccccc|cccccccc@{}}
\toprule
\multirow{2}{*}{IC} & \multicolumn{6}{c|}{Filter size $C_{i}$} & \multicolumn{8}{c}{Depth size $d_{i}$}\\ 
\cmidrule{2-7} 
\cmidrule{2-15}
& $C_{1}$ & $C_{2}$ & $C_{3}$ & $C_{4}$ & $C_{5}$ & $C_{6}$ & $d_{0}$ & $d_{1}$ & $d_{2}$ & $d_{3}$ & $d_{4}$ & $d_{5}$ & $d_{6}$ & $d_{7}$ \\
\midrule
B1  & 320 & 320 & 320 & 320 & 192 & 192 & $-$ & $-$ & $-$ & $-$ & $-$ & $-$ & $-$ & $-$ \\
B2  & 128 & 192 & 256 & 320 & 192 & 192 & $-$ & 2 & 2 & 6 & 2 & 5 & 1 & $-$ \\
O1  & 128 & 192 & 256 & 320 & 192 & 192 & $-$ & 2 & 2 & 6 & 2 & 5 & 1 & 2\\
O2  & 128 & 192 & 256 & 320 & 192 & 192 & 3 & 2 & 2 & 6 & 2 & 5 & 1 & 2\\
\bottomrule
\end{tabular}
\end{table}

\subsection{Experimental Setup}
\label{experiments}
\begin{figure}[htbp!]
  \centering
    \includegraphics[width=0.8\linewidth]{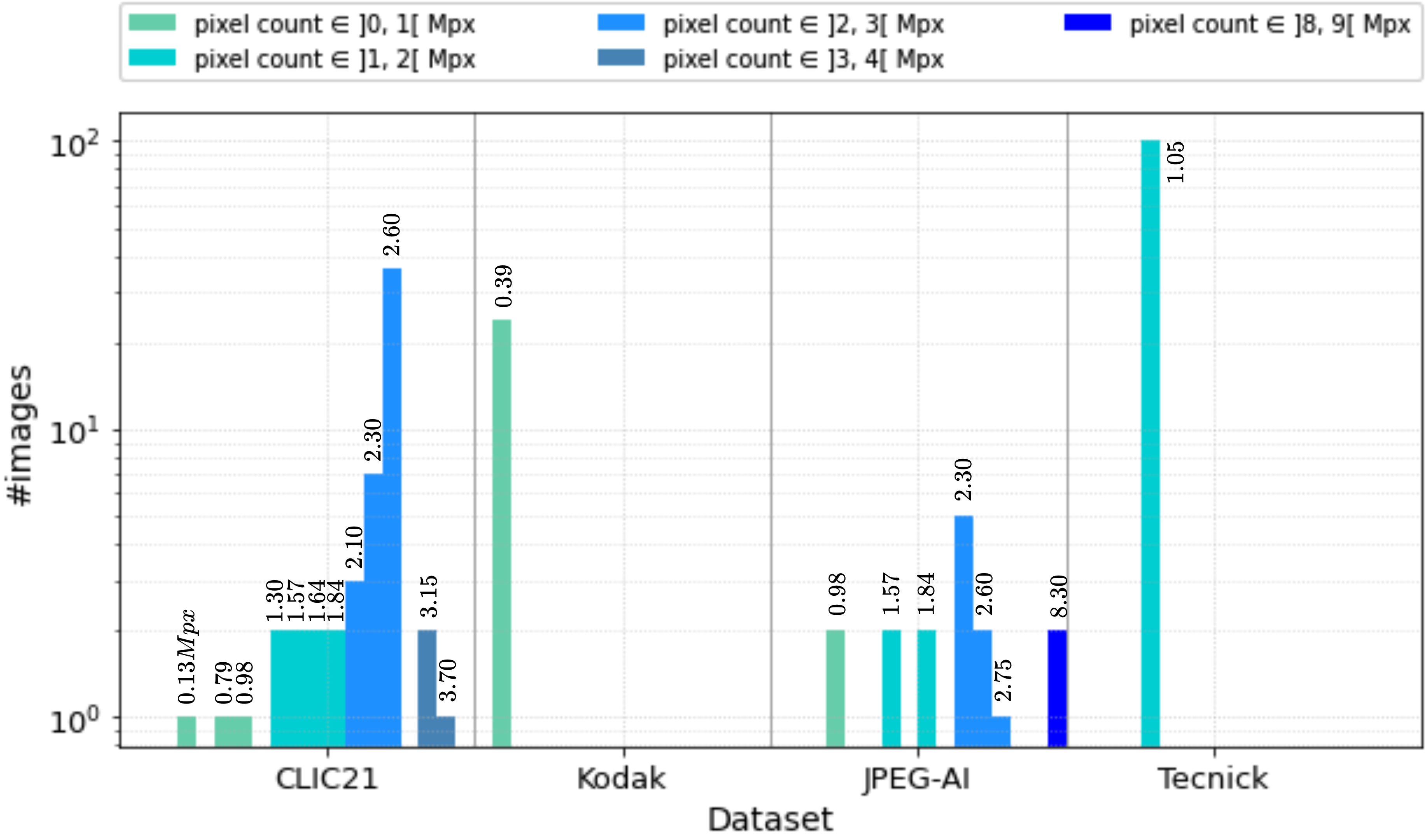}
  \caption{Number of images per dataset per pixel count in megapixel (Mpx).}
  \label{datasets}
\end{figure}

{\bf Datasets.}
The training set of the CLIC2020 dataset is used to train the proposed models. This dataset contains professional and user-generated content images in RGB color and grayscale formats. We evaluate image compression models on four datasets, including Kodak~\cite{kodak}, Tecnick~\cite{asuni2014testimages}, JPEG-AI~\cite{jpegai}, and the testing set of CLIC{21}~\cite{clic21}. Fig.~\ref{datasets} gives the number of images by pixel count for the four test datasets. Finally, for a fair comparison, all images are cropped to the highest possible multiples of 256 to avoid padding for neural codecs.
\\\\
{\bf Baselines.}
We compare our approach with the state-of-art neural compression method SwinT-ChARM proposed by Zhu et al. \cite{zhu2021transformer}, and the Conv-ChARM proposed by Minnen et al. \cite{minnen2020channel}, and non-neural compression methods, including \ac{bpg}(4:4:4), and the up-to-date \ac{vvc} official Test Model VTM-18.0 in All-Intra profile configuration.
Table~\ref{config} gives the configuration of each of the considered image codec baselines with B1 and B2 referring to Conv-ChARM and SwinT-ChARM, respectively, and O1 and O2 refer to our proposed approaches \ac{ict} and \ac{aict}, respectively. $C_{i}$ and $d_{i}$ are the hyperparameters defined in Fig.~\ref{ov_framework}.
We intensively compare our solutions with Conv-ChARM \cite{minnen2020channel} and SwinT-ChARM \cite{zhu2021transformer} from the state-of-the-art models \cite{zou2022devil,zhu2022unified,kim2022joint,he2022elic,el-nouby2023image,muckley2023improving}, under the same training and testing conditions. Nevertheless, Fig.~\ref{bdr_vs_dt} compares our models with additional state-of-the-art solutions.
\\\\
{\bf Implementation details.}
We implemented all models in TensorFlow using \ac{tfc} library \cite{tfc_github}, and the experimental study was carried out on an RTX 5000 Ti \ac{gpu}. All models were trained on the same CLIC2020 training set with 2M steps using the ADAM optimizer with parameters $\beta_1=0.9$ and $\beta_2=0.999$. The initial learning rate is set to $10^{-4}$ and drops to $10^{-5}$ for the last 200k iterations. The loss function, expression in Equation~\eqref{lossfct}, 
is a weighted combination of bitrate $R$ and distortion $D$, with $\lambda$ being the Lagrangian multiplier steering rate-distortion trade-off.  \ac{mse} is used as the distortion metric in RGB color space. Each training batch contains eight random crops $\bm{x}^j \in R^{256 \times 256 \times 3}$ from the CLIC2020 training set. To cover a wide range of rate and distortion points, for our proposed method and respective ablation models, we trained four models with $\lambda \in \{1000, 200, 20, 3\} \times 10^{-5}$. The inference time experiments on the \ac{cpu} are performed on an Intel(R) Xeon(R) W-2145 processor running at 3.70 GHz.

\begin{table}[t]
\centering
\caption{BD-rate$\downarrow$ (\acs{psnr}) performance of \ac{bpg} (4:4:4), Conv-ChARM, SwinT-ChARM, \ac{ict}, and \ac{aict} compared to the VTM-18.0.}\label{bdrate}

\begin{tabular}{@{}l|ccccc@{}}
\toprule
Image Codec  & Kodak & Tecnick & JPEG-AI & CLIC21 & {\bf Average }\\
\midrule
BPG444       & 22.28\%          & 28.02\%	       & 28.37\%          & 28.02\% &      26.67\%   \\
Conv-ChARM   &  2.58\%          &  3.72\%          &  9.66\%          &  2.14\%   &    4.53\%  \\
SwinT-ChARM  & -1.92\%          & -2.50\%          &  2.91\%          & -3.22\%    &  -1.18\%   \\
ICT (ours)   & \textbf{-5.10\%} & -5.91\%	       & -1.14\%          & -6.44\%     &  -4.65\%  \\
AICT (ours)  & 
\textbf{-5.09\%} & \textbf{-5.99\%} & \textbf{-2.03\%} & \textbf{-7.33\%} & \textbf{-5.11\%} \\
\bottomrule
\end{tabular}%
\end{table}

\begin{figure*}[htbp!]
\centering
\includegraphics[width=1\textwidth]{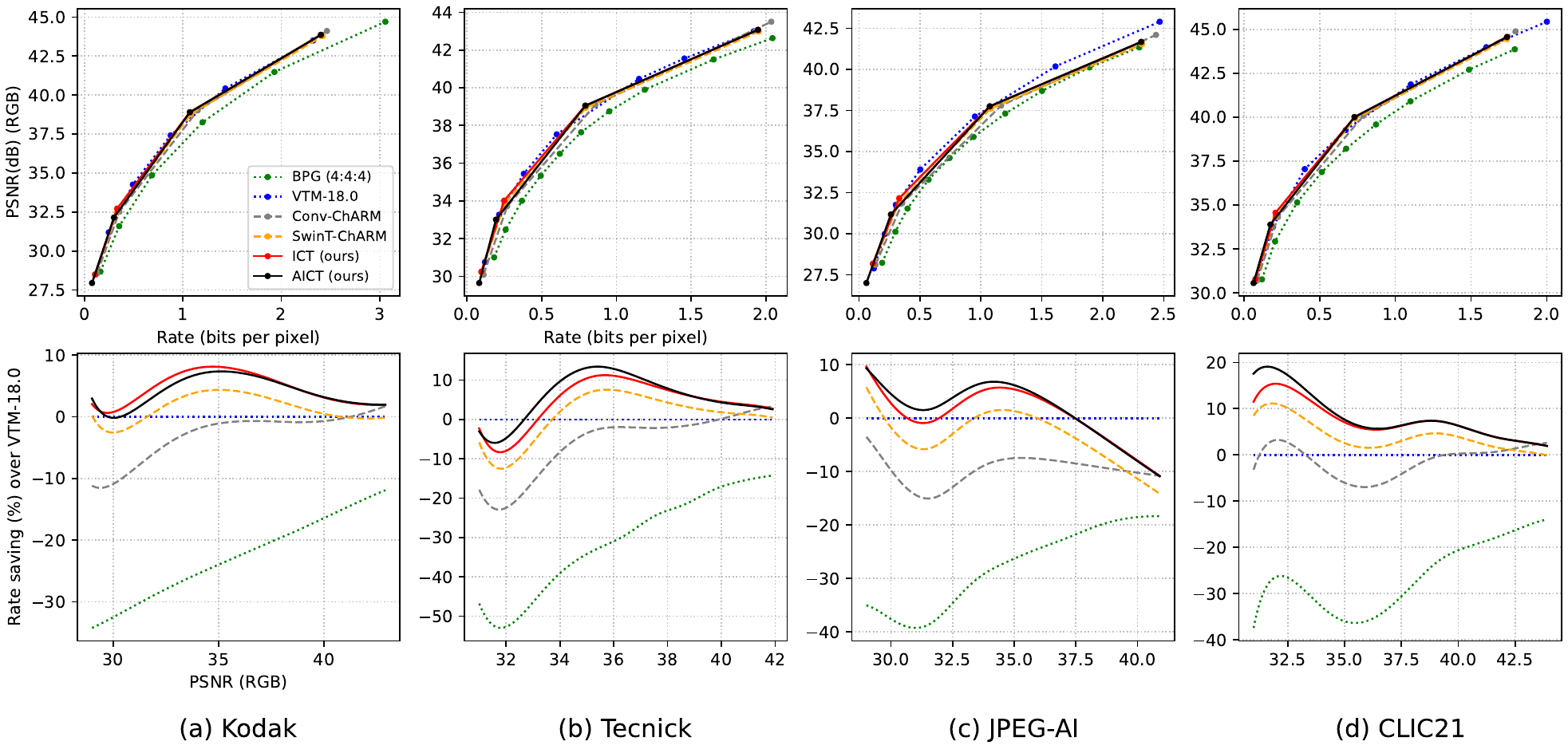}
\caption{Comparison of compression efficiency on Kodak, Tecnick, JPEG-AI, and CLIC21 datasets. Rate-distortion (\acs{psnr} vs. rate (bpp)) comparison and rate saving over VTM-18.0 (larger is better) are respectively illustrated for each benchmark dataset.}
\label{rd_curves}
\end{figure*}
\subsection{Rate-Distortion Performance}
%
To demonstrate the compression efficiency of our proposed solutions, we plot the rate-distortion curves of \ac{ict}, \ac{aict}, and the baselines on benchmark datasets. Fig.~\ref {rd_curves} (a) ($1^{st}$ row) gives the \acs{psnr} versus the bitrate for our solutions and baselines on the Kodak dataset. The latter figure shows that \ac{aict} and \ac{ict} equally outperform the neural approaches Conv-ChARM and SwinT-ChARM, as well as \ac{bpg}(4:4:4) and VTM-18.0 traditional codecs, achieving a higher \acs{psnr} values for the different bitrate ranges.

Moreover, we introduce Fig.~\ref{rd_curves} (a) ($2^{nd}$ row), showing the rate savings over the VTM-18.0 on the Kodak dataset.
\textcolor{black}{The rate saving over \ac{vvc} (\%) represents the percentage reduction in bitrate achieved by a specific compression model compared to a reference codec while maintaining an equivalent level of image reconstruction quality, as measured by \ac{psnr} in our context. This graph is a generalized version of a Bjøntegaard Delta (BD) chart \cite{bjontegaard2001calculation} by plotting rate savings as a function of quality, instead of solely presenting average savings \cite{minnen2020channel}. By comparing the performance of different models using this figure, we can discern which model excels in striking a balance between compression efficiency at different levels of reconstruction quality. Models that achieve higher bitrate savings over \ac{vvc} at various \ac{psnr} levels are considered superior in terms of compression performance.}
\ac{ict} and \ac{aict} achieve significant rate savings compared to the baselines, demonstrating their ability to compress images more efficiently. More specifically, \ac{aict}, including the adaptive resolution module, achieves the highest bitrate gain at a low bitrate/quality range, where it is more beneficial to reduce the spatial resolution.  
%
To further generalize the effectiveness of our solutions, we extend the evaluation to three high resolutions datasets (Tecnick, JPEG-AI, and CLIC21), as shown in Fig.~\ref{rd_curves}. The figure illustrates the \acs{psnr} versus bitrate ($1^{st}$ row), rate savings ($2^{nd}$ row) on the considered datasets. \ac{aict} and \ac{ict} consistently achieve better rate-distortion performance and considerable rate savings compared to the existing traditional codecs and the neural codecs Conv-ChARM and SwinT-ChARM, demonstrating their efficiency across different high-resolution images and datasets.
\begin{table}[t]
\centering
\caption{BD-rate$\downarrow$ performance of SwinT-ChARM, \ac{ict}, and \ac{aict} compared to the Conv-ChARM.}\label{bdrate_psnr_msssim}
\begin{tabular}{@{}l|ccccc@{}}
\toprule
Image Codec  & Kodak & Tecnick & JPEG-AI & CLIC21 & {\bf Average }\\
\midrule
  & \multicolumn{5}{c}{BD-rate (\acs{psnr})$\downarrow$}           \\
\midrule
SwinT-ChARM  & -4.24\%           & -6.40\%          & -6.13\%          & -5.37\%          & -5.54\%           \\
ICT (ours)   & \textbf{-7.30\%}  & \textbf{-9.52\%} & -9.85\%          & -8.47\%          & -8.79\%           \\
AICT (ours)  & -7.28\%           & \textbf{-9.68\%} & \textbf{-10.20\%} & \textbf{-9.35\%} & \textbf{-9.13\%} \\
\midrule
  & \multicolumn{5}{c}{BD-rate (\acs{ms-ssim})$\downarrow$}           \\
\midrule
SwinT-ChARM  & -6.34\%           & -7.01\%          & -7.49\%          & -6.30\%          & -6.79\%          \\
ICT (ours)   & \textbf{-7.60\%}  & \textbf{-8.31\%} & -9.29\%          & -7.50\%          & -8.18\%          \\
AICT (ours)  & -7.58\%           & \textbf{-8.31\%} & \textbf{-9.87\%} & \textbf{-7.67\%} & \textbf{-8.36\%} \\
\bottomrule
\end{tabular}%
\end{table}

\begin{figure*}[htbp!]
\centering
\includegraphics[width=1\textwidth]{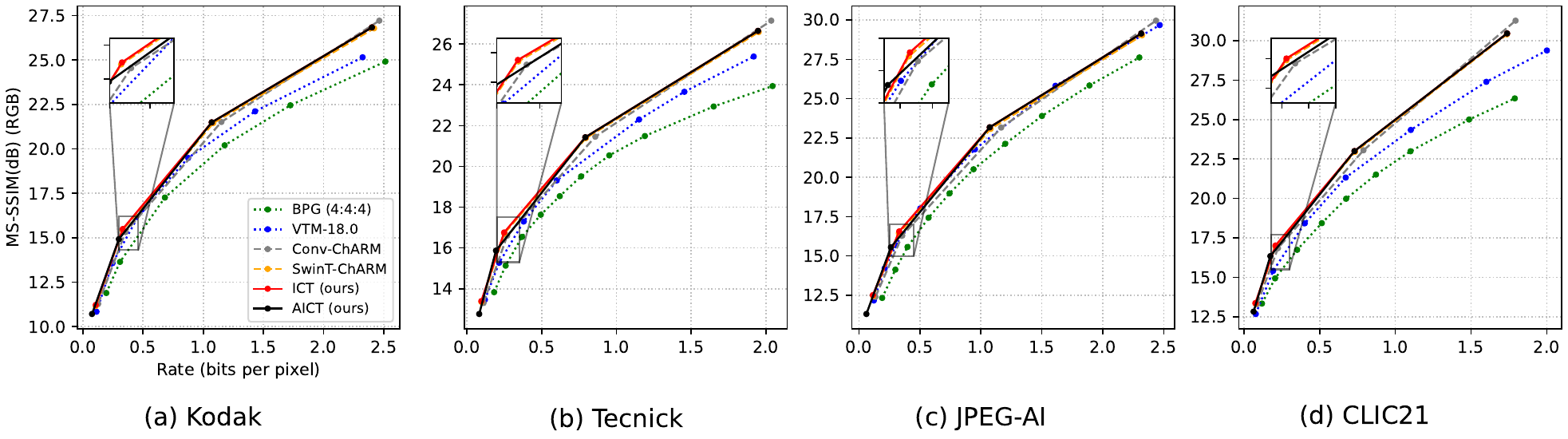}
\caption{{\color{black}Comparison of compression efficiency on the Kodak, Tecnick, JPEG-AI, and CLIC21 datasets. Rate distortion (\acs{ms-ssim} vs the rate (bpp)) comparison is illustrated for each benchmark dataset.}}
\label{rd_curves_msssim}
\end{figure*}

Furthermore, we assessed the effectiveness of our methods using the perceptual quality metric \ac{ms-ssim} on the four benchmark datasets.
\textcolor{black}{To calculate \ac{ms-ssim} in decibels (dB), a logarithmic transformation is applied to the original MS-SSIM values as follows: $MSSSIM (dB) = -10 \times \log_{10}(1 - MSSSIM)$. This transformation is performed to provide a more intuitive and interpretable scale for comparing image quality, as previously done in several works \cite{he2022elic,el-nouby2023image,kim2022joint,10.1145/3474085.3475213,zhu2022unified}.}
Fig.~\ref{rd_curves_msssim} gives the \ac{ms-ssim} scores expressed versus the bitrate for the four test datasets. As illustrated in Fig.~\ref{rd_curves_msssim}, our methods yield better coding performance than the current neural baselines in terms of \ac{ms-ssim}. Note that we haven't optimized our approaches and baselines using \ac{ms-ssim}, as a differentiable distortion measure in the loss function, during the training process. Thus, optimizing the solutions with \ac{ms-ssim} will further improve the performance regarding this metric.

Besides the rate-distortion rate savings curves, we also evaluate different models using Bjontegaard's metric \cite{bjontegaard2001calculation}, which computes the average bitrate savings (\%) between two rate-distortion curves. In Table~\ref{bdrate}, we summarize the BD-rate (\acs{psnr}) of image codecs across all four datasets, compared to the VTM-18.0 as the anchor. On average, \ac{ict} and \ac{aict} are able to respectively achieve -4.65\% and -5.11\% rate reductions compared to VTM-18.0 and -3.47\% and -3.93\% relative gain from SwinT-ChARM.
In addition, Table~\ref{bdrate_psnr_msssim} presents the BD-rate (\acs{psnr}/\acs{ms-ssim}) of SwinT-ChARM, \ac{ict}, and \ac{aict} across the considered datasets, compared with the anchor Conv-ChARM. Once again, \ac{ict} and \ac{aict} are able to outperform the neural approach Conv-ChARM, with average rate reductions (\acs{psnr}) of -8.79\% and -9.13\% and average rate reductions (\acs{ms-ssim}) of -8.18\% and -8.36\%, respectively, outperforming SwinT-Charm solution on the four benchmark datasets and the two image quality metrics. 

Overall, the proposed \ac{ict} and \ac{aict} have demonstrated strong rate-distortion performance on various benchmark datasets. This indicates that our approaches can better preserve image quality at lower bitrates, highlighting its potential for practical applications in image compression.

\subsection{Models Scaling Study}
We evaluated the decoding complexity of the proposed and baseline neural codecs by averaging decoding time across 7000 images at 256$\times$256$\times$3 resolution, encoded at varying bitrates, specifically \{0.1, 0.8, 1.5\} (bpp). Subsequently, we computed the average decoding time for this dataset, resulting in an overall average bit rate of 0.8bpp. Table~\ref{complexity} gives the image codec complexity features, including the decoding time on \ac{gpu} and \ac{cpu}, \ac{flops}, and the total model parameters. Finally, we recall that the models run with Tensorflow 2.8 on a workstation with one RTX 5000 Ti \ac{gpu}. 
\begin{figure}[t]
\centering
\includegraphics[width=0.6\linewidth]{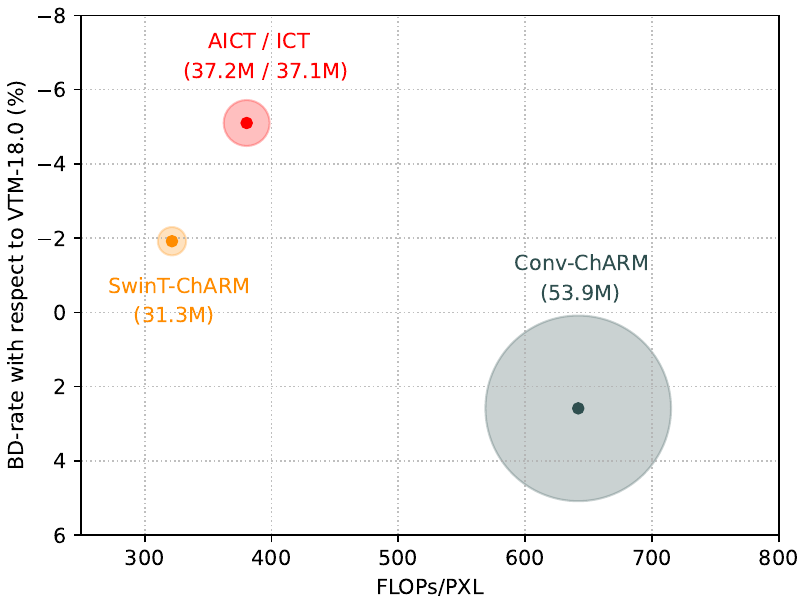}
\caption{Model size scaling. BD-rate (\acs{psnr}) on Kodak dataset versus \ac{flops} per pixel for the proposed \ac{aict} and \ac{ict} compared to Conv-ChARM and SwinT-ChARM (for both encoding and decoding). Circle sizes indicate the numbers of parameters. Left-top is better. }
\label{bdr_vs_flops}
\end{figure}

Compared to the neural baselines, \ac{ict} can achieve faster decoding speed on \ac{gpu} but not on \ac{cpu}, which proves the parallel processing ability to speed up compression on \ac{gpu} and the well-engineered designs of both transform and entropy coding. This is potentially helpful for conducting high-quality real-time visual data streaming. Our \ac{aict} is on par with \ac{ict} in terms of the number of parameters, \ac{flops}, and latency, indicating the lightweight nature of the scale adaptation module with consistent coding gains over four datasets and two quality metrics.

Fig.~\ref{bdr_vs_flops} gives the BD-rate (with VTM-18.0 as an anchor) performance versus the \ac{flops} per pixel of the \ac{ict}, \ac{aict}, SwinT-ChARM and Conv-ChARM on the Kodak dataset. We can notice that \ac{ict}  and \ac{aict} are in an interesting area, achieving a good trade-off between BD-rate score on Kodak, total model parameters, and \ac{flops} per pixel, reflecting an efficient and hardware-friendly compression model.

\begin{figure}[t]
\centering
\includegraphics[width=0.7\linewidth]{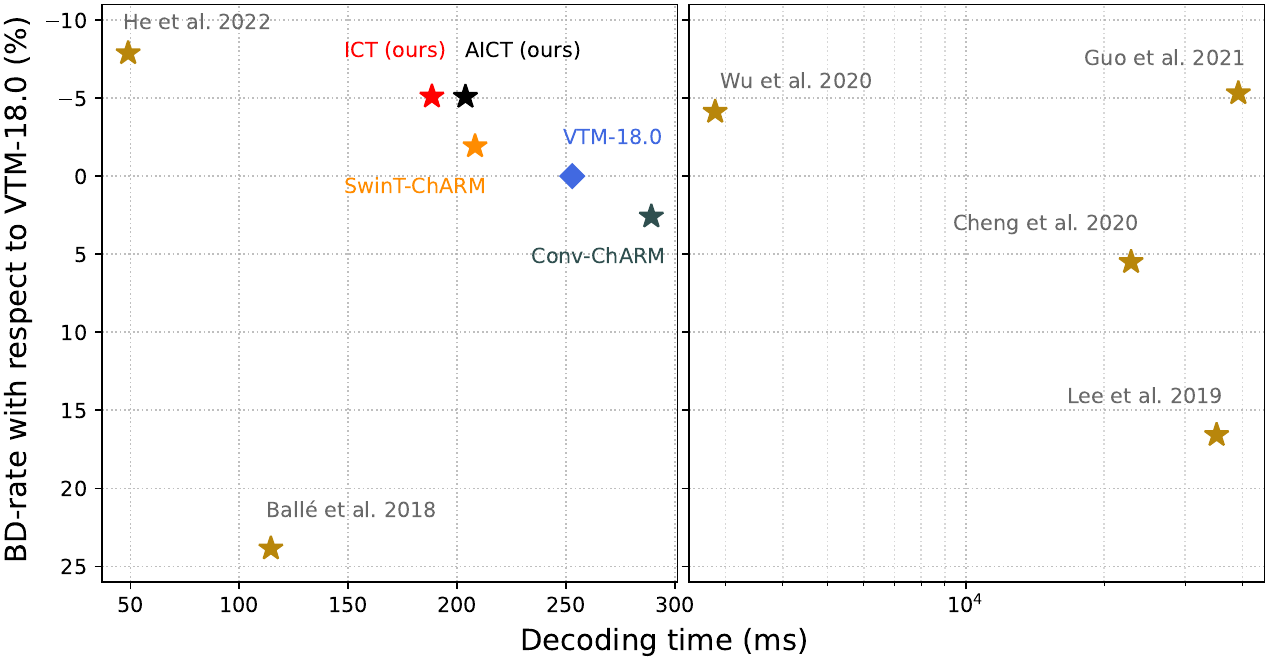}
\caption{BD-rate (\acs{psnr}) versus decoding time (ms) on the Kodak dataset. Star and diamond markers refer to decoding on \ac{gpu} and \ac{cpu}, respectively. Left-top is better.}
\label{bdr_vs_dt}
\end{figure}

Finally, Fig.~\ref{bdr_vs_dt} shows the BD-rate (with VTM-18.0 as an anchor) versus the decoding time of various codecs on the Kodak dataset. It can be seen from the figure that our \ac{ict} and \ac{aict} achieve a good trade-off between BD-rate performance and decoding time. Furthermore, the symmetrical architecture of the proposed solutions allows similar complexity at both the encoder and decoder. This feature can be an advantage of neural codecs, since the best conventional codecs like \ac{vvc} exhibit more complex encoding than decoding.

\begin{table}[!t]
\centering
\caption{Image codec complexity. We calculated the average decoding latency across 7000 images at 256$\times$256 resolution, encoded on average at 0.8 bpp. The best score is highlighted in bold.}\label{complexity}
\begin{tabular}{@{}l|cc|c|c@{}}
\toprule
\multirow{2}{*}{Image Codec} & \multicolumn{2}{c|}{{Latency(ms)$\downarrow$}} & \multirow{2}{*}{M\ac{flops}$\downarrow$} & \multirow{2}{*}{\#parameters (M)$\downarrow$}   \\ \cmidrule{2-3}
& \ac{gpu}  & \ac{cpu} & & \\
\midrule
Conv-ChARM   & 133.8         & \textbf{359.8} & 126.1999 & 53.8769 \\
SwinT-ChARM  & 91.8          & 430.7          & 63.2143  & 31.3299 \\
ICT (ours)   & \textbf{80.1} & 477.0          & 74.7941  & 37.1324 \\
AICT (ours)  & 88.3          & 493.3          & 74.9485  & 37.2304 \\
\bottomrule
\end{tabular}%
\end{table}

\subsection{Resize Parameter Analysis}
We conduct a resize parameter analysis through the benchmark datasets, including the images of the four datasets with various resolutions, as illustrated in Fig.~\ref{datasets}. 

\begin{figure}[t]
  \centering
    \includegraphics[width=0.65\linewidth]{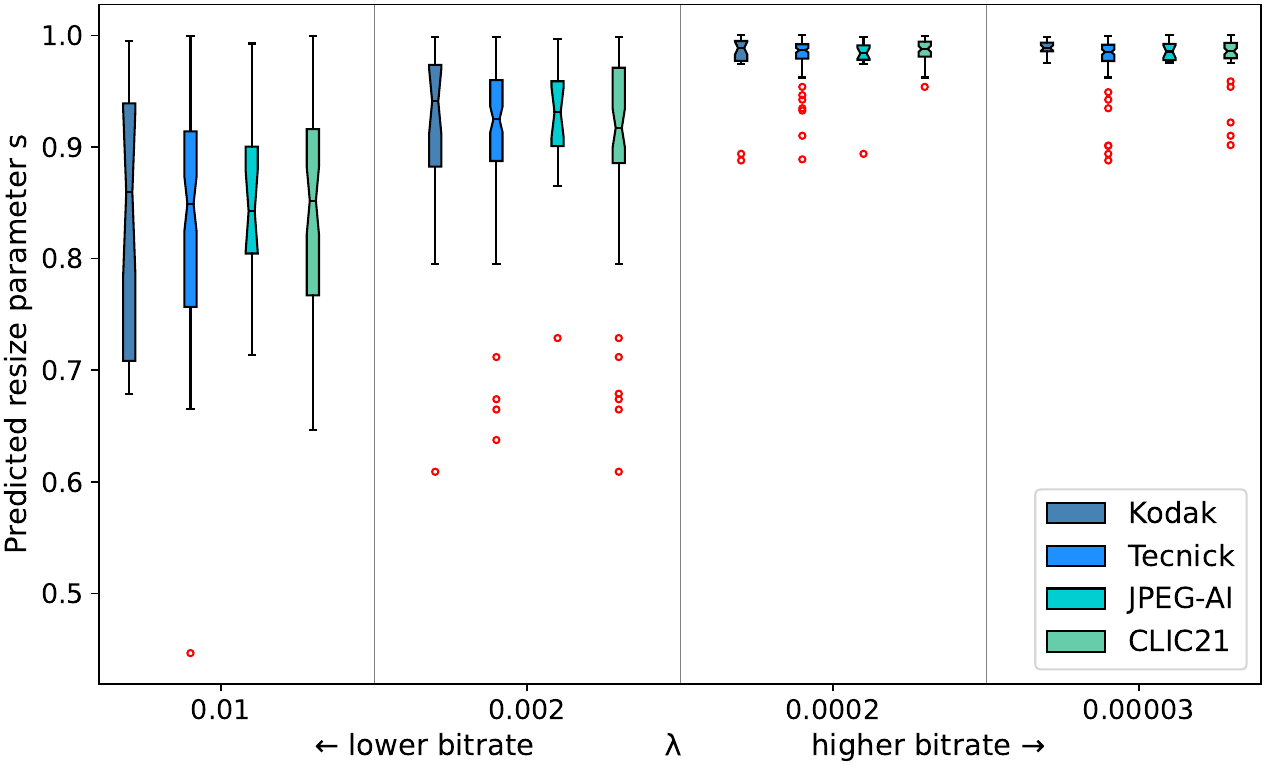}
  \caption{Box plot of predicted resize parameter $s$ versus the weighting parameter $\lambda$, evaluated across the four considered datasets. The '$\circ$' symbol denotes outliers.}
  \label{resize_param}
\end{figure}

Fig.~\ref{resize_param} shows how the parameter $s$ varies according to the weighting parameter $\lambda$ (i.e., bitrate) for the four datasets. First, we can notice that the estimated resize parameter $s$ depends on the bitrate and the spatial characteristics of the image content. Resizing the input image to a lower resolution is frequently observed at a low bitrate, where the compression removes the image details. In contrast, the down-sampling is not performed at a high bitrate to reach high image quality, significantly when the up-sampling module cannot recover the image details at the decoder. Nevertheless, even at a high bitrate, a few samples are down-sampled to a lower resolution, especially images with low spatial information that the up-sampling module can easily recover on the decoder side. This also can explain the higher coding gain brought by the adaptive sampling module of \ac{aict} on datasets, including more high-resolution images such as JPEG-AI and CLIC21 (see Fig.~\ref{datasets}).
\textcolor{black}{To gain a deeper understanding of the observed performance variations in Tables~\ref{bdrate} and \ref{bdrate_psnr_msssim}, we direct our focus towards the characteristics of the test datasets. Illustrated in Fig.~\ref{datasets}, the distribution of images by pixel count for the four test datasets, including Kodak, Tecnick, and two datasets (JPEG-AI and CLIC21) with higher-resolution images. It is noteworthy that, as depicted in the Fig.~\ref{datasets}, Kodak and Tecnick comprise images with a significantly lower total number of pixels. This particular attribute inherently impacts the effectiveness of certain modules, like adaptive scaling, given the correlation between the predicted resize factor and the total number of pixels in each image. Furthermore, the content complexity within these datasets may contribute to the observed limited impact of adaptive scaling. As demonstrated in the Fig.~\ref{resize_param}, images with higher high-frequency details might not experience as significant benefits from such modules.}
In addition, skipping the resize modules for a predicted scale close to 1 $s \cong 1$ contribute to reducing encoding and decoding complexity.

\begin{figure*}[htbp!]
  \centering
  \subfloat[Conv-ChARM]{
    \includegraphics[width=0.31\linewidth]{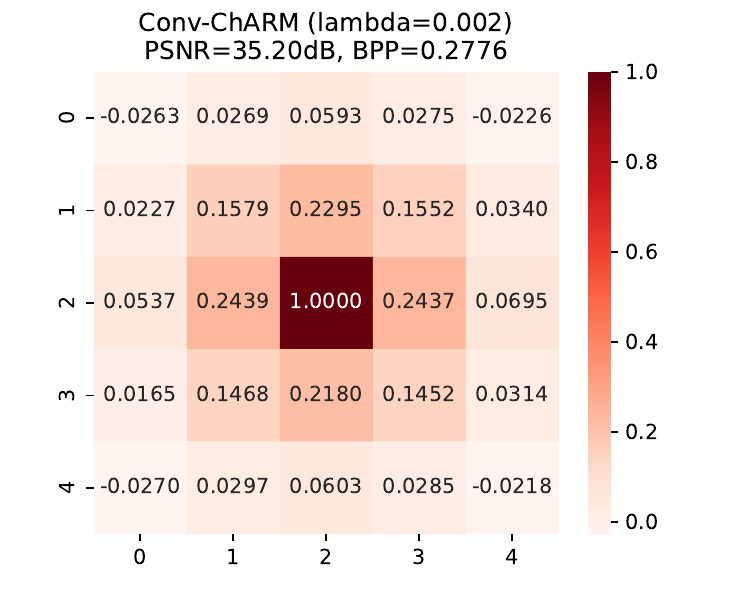}
    \label{ms_latent}}
  \hfil
  \subfloat[SwinT-ChARM]{
    \includegraphics[width=0.31\linewidth]{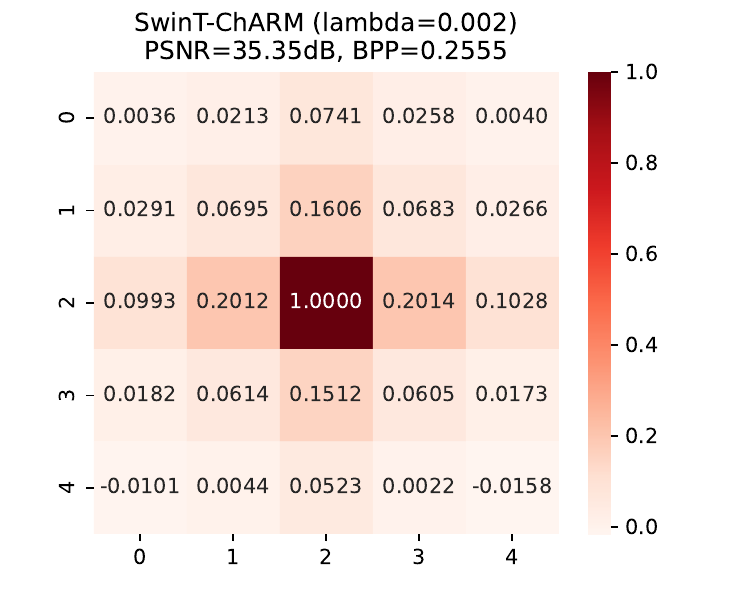}
    \label{swin_latent}}
  \hfil
  \subfloat[\acs{ict}]{
    \includegraphics[width=0.31\linewidth]{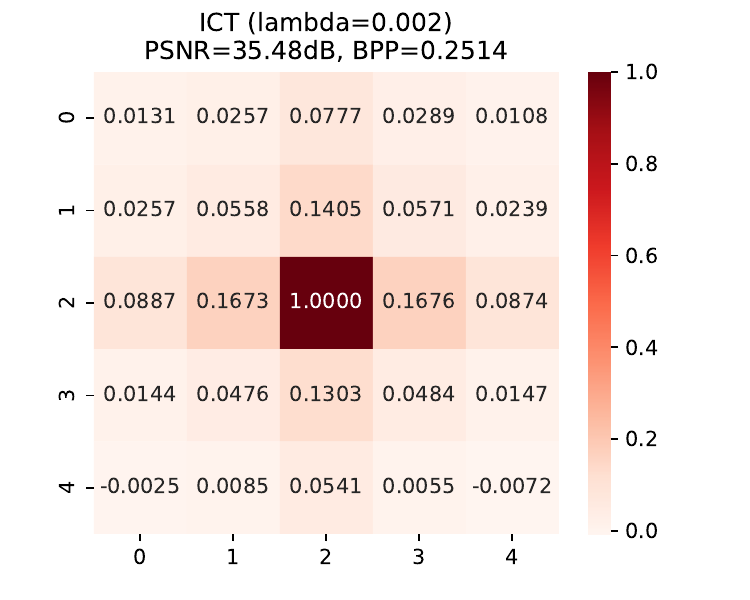}
    \label{ict_latent}}
  \caption{The spatial correlation at index (i, j) corresponds to the normalized cross-correlation of the reconstructed latent $\frac{(\hat{y} - \mu)}{\sigma}$ at spatial location $( w_{c}, h_{c} )$ and $( w_{c} + i, h_{c} + j )$, averaged across all latent channels of all image patches across the four considered datasets. We considered (a) Conv-ChARM, (b) SwinT-ChARM, and the proposed (c) \ac{ict}, all trained at $\lambda=0.002$.}
  \label{latent_analysis}
\end{figure*}

\begin{figure*}[htb]
\centering
\includegraphics[width=1\textwidth]{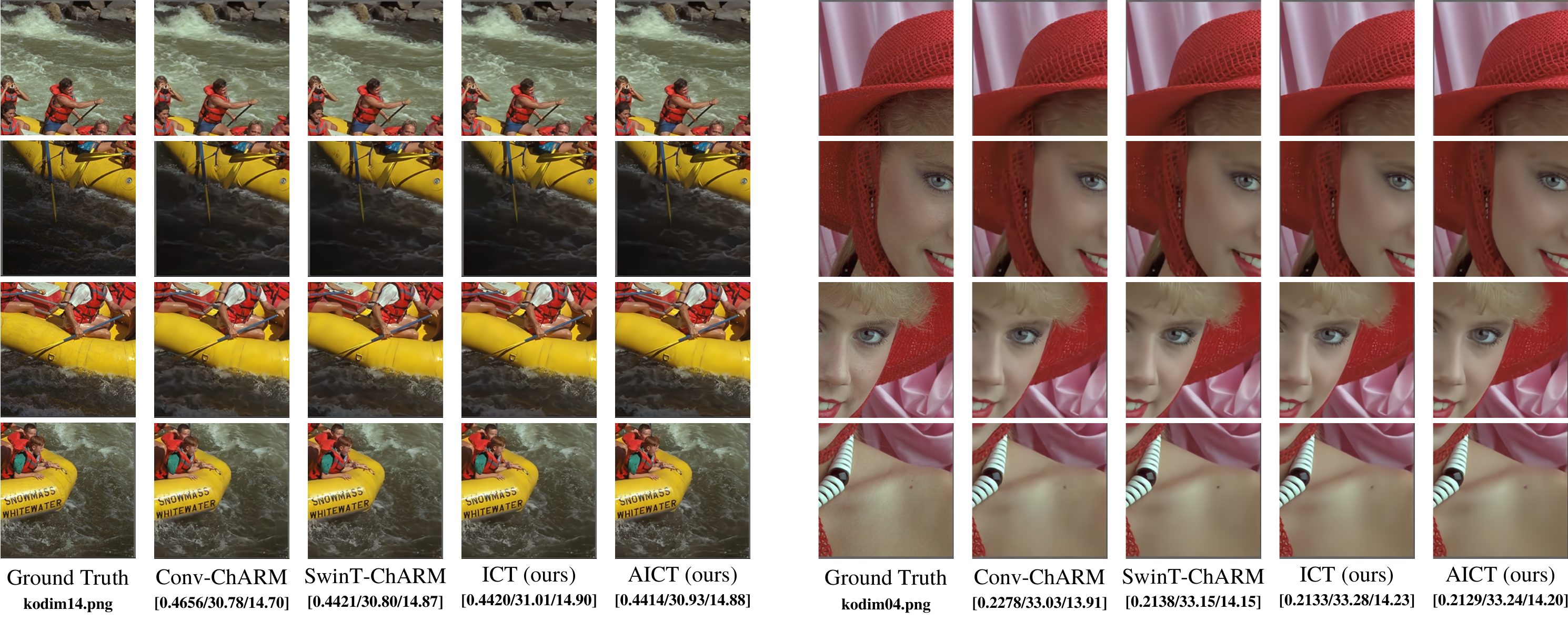}
\caption{Visualization of the reconstructed images from the Kodak dataset. The metrics are [bpp↓/\acs{psnr}(dB)↑/\acs{ms-ssim}(dB)↑].}
\label{qual_kodak}
\end{figure*}

\subsection{Latent Analysis}
\label{latent}
Transform coding is motivated by the idea that coding is more effective in the transform domain than in the original signal space. A desirable transform would decorrelate the source signal so that a simple scalar quantization and factorized entropy model can be applied without constraining coding performance. Furthermore, an appropriate prior model would provide context adaptivity and utilize distant spatial relations in the latent tensor.
The effectiveness of the analysis transforms $g_{a}$ can then be evaluated by measuring the level of correlation in the latent signal $\hat{\bm{y}}$. We are particularly interested in measuring the correlation between nearby spatial positions, which are heavily correlated in the source domain for natural images. In Fig.~\ref{latent_analysis}, we visualize the normalized spatial correlation of $\hat{\bm{y}}$ averaged over all latent channels and compare Conv-ChARM and SwinT-ChARM with the proposed \ac{ict} at $\lambda=0.002$.
We can observe that while both lead to small cross-correlations, \ac{ict} can decorrelate the latent with a slight improvement when compared to SwinT-ChARM and a much considerable improvement when compared to Conv-ChARM. This suggests that Transformer-based transforms with Transformer-based entropy modeling incur less redundancy across different spatial latent locations than convolutional ones, leading to an overall better rate-distortion trade-off.

\subsection{Ablation Study}
\label{ablation}
%
To investigate the impact of the proposed \ac{ict} and \ac{aict}, we conduct an ablation study according to the reported BD-rate$\downarrow$ w.r.t. \ac{vvc} and Conv-ChARM in Table~\ref{bdrate} and Table~\ref{bdrate_psnr_msssim}, respectively. 
Image compression performance increases from Conv-ChARM to SwinT-ChARM on the considered datasets due to the inter-layer feature propagation across non-overlapping windows (local information) and self-attention mechanism (long-range dependencies) in the Swin Transformer.
With the proposed spatio-channel entropy model, \ac{ict} is able to achieve on average -3.47\% (\acs{psnr}) and -1.39\% (\acs{ms-ssim}) rate reductions compared to SwinT-ChARM.
Moreover, \ac{aict} is enhancing \ac{ict} on average by -0.46\% (\acs{psnr}) and -0.18\% (\acs{ms-ssim}) rate reductions with consistent gain over the four datasets. This indicates that introducing a scale adaptation module can further reduce spatial redundancies and alleviate coding artifacts, especially at low bitrate for higher compression efficiency. More importantly, the adaptive resolution may also reduce the complexity of the encoder and decoder regarding the number of operations per pixel since fewer pixels are processed on average by the codec when the input image is downscaled to a lower resolution, i.e., $s < 1$.  

\subsection{Qualitative Analysis}
To assess the perceptual quality of the decoded images, we visualize two reconstructed samples with the proposed \ac{ict} and \ac{aict} methods, along with Conv-ChARM and SwinT-ChARM, all trained at the same low bitrate configuration ($\lambda=0.002$). Fig.~\ref{qual_kodak} presents the visualization of the reconstructed {\it kodim14} and {\it kodim04 } images from the Kodak dataset.
\textcolor{black}{Although not immediately apparent in Fig.~\ref{qual_kodak}, the proposed \ac{ict} and \ac{aict} may manifest subtle improvements in specific image regions or content types. A closer examination reveals discernible differences in certain areas of the reconstructed images. For instance, in the 'kodim14.png' figure, there is a noticeable enhancement in the intensity of black pixels within the text elements in the proposed models, \ac{ict} and \ac{aict}, compared to the baselines. This signifies a finer level of detail preservation in the text, which can be particularly crucial in applications involving textual content. Moreover, in the same 'kodim14.png' figure, in the last row of patches, it can be observed that the representation of water soaking the green sweater worn by the person is significantly better in the \ac{ict} model compared to the other models. This improved depiction of intricate textures and finer details highlights the model's capability to faithfully capture and reproduce complex image features, which may be essential in scenarios where preserving fine textures is critical for maintaining perceptual quality.}
In summary, under a similar rate budget, \ac{ict} and \ac{aict} perform better in maintaining texture details and clean edges while suppressing visual artifacts of the decoded images compared to Conv-ChARM and SwinT-ChARM neural approaches. Additionally, the self-attention mechanism focuses more on high-contrast image regions and consequently achieves more coding efficiency on such content.

%
\section{Conclusion}
\label{concl}
In this work, we have proposed an up-and-coming neural codec \ac{aict}, achieving compelling rate-distortion performance while significantly reducing the latency, which is potentially helpful to conduct, with further optimizations, high-quality real-time visual data transmission. 
We inherited the advantages of self-attention units from Transformers to approximate the mean and standard deviation for efficient entropy modeling and combine global and local texture to capture correlations among spatially neighboring elements for non-linear transform coding.
Furthermore, we have presented a lightweight spatial resolution scale adaptation module to enhance compression ability, especially at low bitrates.
The experimental results, conducted on four datasets, showed that \ac{ict} and \ac{aict} approaches outperform the state-of-the-art conventional codec \ac{vvc}, achieving respectively -4.65\% and -5.11\% BD-rate reductions compared to the VTM-18.0 by averaging over the benchmark datasets.
With the development of \ac{gpu} and \ac{npu} chip technologies and further universal optimization frameworks \cite{10.1145/3580499}, neural codecs will be the future of visual data coding, achieving better compression efficiency when compared with traditional codecs and aiming to bridge the gap to a real-time processing.


\bibliographystyle{ACM-Reference-Format}
\bibliography{ref}

\end{document}